\pdfoutput=1
\documentclass[11pt]{article}

\usepackage[final]{acl}

\usepackage{times}
\usepackage{latexsym}

\usepackage[T1]{fontenc}
\usepackage[utf8]{inputenc}

\usepackage{microtype}
\usepackage{enumitem}% http://ctan.org/pkg/enumitem

\usepackage{inconsolata}

\usepackage{graphicx}
\usepackage{xcolor}
\usepackage{amsmath}
\usepackage{subcaption}
\usepackage{float}
\usepackage{tikz}
\usepackage{booktabs}
\usepackage{adjustbox}
\usepackage{placeins}
\usepackage{duckuments}

\title{Signature vs. Substance: Evaluating the Balance of Adversarial Resistance and Linguistic Quality in Watermarking Large Language Models}

\author{William Guo \\
  IMSA, \\
  Aurora, Illinois, USA\\
  \texttt{wguo@imsa.edu} \\\And
  Ana Smith \\
  MIT Lincoln Laboratory,\\
  Lexington, MA, USA \\
  \texttt{ana.smith@ll.mit.edu} \\ \And
  Adaku Uchendu \\
  MIT Lincoln Laboratory,\\
  Lexington, MA, USA \\
  \texttt{adaku.uchendu@ll.mit.edu}
  }

\begin{document}
\maketitle
\begin{abstract}
To mitigate the potential harms of Large Language Models (LLMs)
generated text, researchers have proposed watermarking, a process of embedding detectable signals within text. 
With watermarking, we can always accurately detect LLM-generated texts. However, recent findings suggest that these techniques often negatively affect the quality of the generated texts, 
and adversarial attacks can strip the watermarking signals, causing the texts to possibly evade detection. 
These findings have created resistance in the wide adoption of watermarking by LLM creators. 
Finally, to encourage adoption, we evaluate the robustness of several watermarking techniques to adversarial attacks by comparing paraphrasing and back translation (i.e., English $\to$ another language $\to$ English) attacks; and 
their ability to preserve quality and writing style of the unwatermarked texts by using linguistic metrics to capture quality and writing style of texts. 
Our results suggest that these watermarking techniques preserve semantics, deviate from the writing style of 
the unwatermarked texts, and are susceptible to adversarial attacks, especially for the back translation attack.

\footnote{DISTRIBUTION STATEMENT A. Approved for public release. Distribution is unlimited.
This material is based upon work supported by the Department of the Air Force under Air Force Contract No. FA8702-15-D-0001. Any opinions, findings, conclusions or recommendations expressed in this material are those of the author(s) and do not necessarily reflect the views of the Department of the Air Force.
© 2024 Massachusetts Institute of Technology.
Delivered to the U.S. Government with Unlimited Rights, as defined in DFARS Part 252.227-7013 or 7014 (Feb 2014). Notwithstanding any copyright notice, U.S. Government rights in this work are defined by DFARS 252.227-7013 or DFARS 252.227-7014 as detailed above. Use of this work other than as specifically authorized by the U.S. Government may violate any copyrights that exist in this work.}

\end{abstract}

\section{Introduction}
% Motivation 
% why should we care
% A brief description of what the problem is and what you did 
% Contributions summary 

Large Language Models (LLMs) have made remarkable improvements, showcasing exceptional abilities in comprehension of natural language, creation of text, code creation, etc. However, the potential misuse of LLMs poses many threats, including the spread of deepfake audio and video, the increase of fake news, and the presence of bias and toxicity in generated content. As these models continue to advance, the proliferation of machine-generated text in particular raises text copyright issues. As a result, the ability to detect and tag LLM-generated text is becoming increasingly important. However, due to the ability of LLMs to mimic humans well, distinguishing machine-generated texts from human texts is extremely difficulty. Therefore, researchers have proposed two techniques - detection \citep{jawahar2020automaticdetectionmachinegenerated} and watermarking \citep{kirchenbauer2024watermarklargelanguagemodels}. We focus on Watermarking because of the increasing prevalence of AI-generated texts being misconstrued as human-written. 
Therefore, in the near future, detectors will only perform at a chance level and with watermarking, machine-generated texts detection will be more possible at a higher certainty \cite{chakraborty2024position}.  

\begin{figure}
    \centering
    \includegraphics[width=0.9\linewidth]{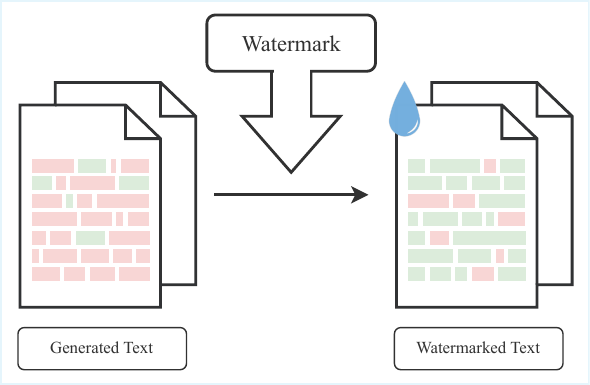}%{example-image-duck}
    \caption{Illustration of the Watermarking process for Large Language Models}
    \label{fig:water-teaser}
\end{figure}

Text watermarking refers to embedding markers in textual content \citep{8268096}. 
Which implies that a detector that is aware of the watermarking technique is able to recognize with high accuracy if the text is produced by an LLM. For example, the watermarking algorithm proposed by \citet{kirchenbauer2024watermarklargelanguagemodels} 
divides the model’s vocabulary into two lists, 
green and red, and boosts the logit values for tokens in the green list by a fixed constant, known as the watermark strength. This adjustment biases the model toward selecting green tokens during text generation, so that the resulting content statistically favors this subset. 
Thus, the watermarked text's prevalence of green tokens, then provides an indicator that the material comes from an LLM, allowing detectors to distinguish it from human-written texts.
See Figure \ref{fig:water-teaser} for an illustration of watermarking for LLMs.

However, while the watermarking technique is a more promising 
way to achieve accurate machine-generated text detection, researchers \citet{sadasivan2025aigeneratedtextreliablydetected} have found two major limitations - 
reduction of quality of generated texts and lack of robustness to adversarial attacks. 
% We investigate these limitations as 
% an ideal watermarking method should not compromise the quality of the output text or negatively impact the language model's usual usage. Moreover, it should exhibit robustness against adversarial attacks like paraphrasing. However, with the rising number of watermarking algorithms, their complex mechanisms, and the diversity of evaluation perspectives and metrics, necessitate the need to make these techniques easy to test, understand, and evaluate. Despite these advances, there remains a notable gap in the linguistic evaluation of watermarking methods. For example, Water Bench\citep{tu2024waterbenchholisticevaluationwatermarks} focuses on holistically evaluating various LLM watermarking under a unified metric; Mark My Words \citep{piet2024markwordsanalyzingevaluating} focuses on the quality, detection efficiency, and tamper resistance of LLM watermarking techniques; and MarkLLM \citep{pan2024markllmopensourcetoolkitllm} focuses on creating a platform for comparing watermark detectability, basic robustness, and text quality. Addressing this gap is critical to ensuring that watermarking not only maintains the technical robustness of LLM outputs but also preserves the natural and intended flow of language, thereby supporting both the functional and aesthetic quality of generated content.
Thus, we comprehensively evaluate the utility of watermarking techniques in the defense against machine-generated text pollution by assessing both quality of watermarked vs. unwatermarked texts, and the robustness of the watermarked texts to techniques that strip the text of its watermarking signatures.
% To bridge this gap, we conducted a comprehensive evaluation of LLM watermarking techniques to improve AI transparency and responsibility. 
This yielded three research questions (RQs):
\begin{itemize}[leftmargin=3.5em,noitemsep]
    \item[\textbf{RQ1}:] How robust are Watermarking techniques to Paraphrasing adversarial attacks?
    \item[\textbf{RQ2}:] How well does Watermarking techniques preserve text quality and writing style?
    \item[\textbf{RQ3}:] Can Linguistic features help us understand why the watermarking techniques are robust or not-robust?
\end{itemize}

First, we select and evaluate several watermarking algorithms, such as 
KGW \citep{kirchenbauer2024watermarklargelanguagemodels}, 
Semantic Invariant Robust Watermark (SIR) \citep{liu2024semanticinvariantrobustwatermark}, 
Entropy-based Text Watermarking (EWD) \citep{lu2024entropybasedtextwatermarkingdetection}, and 
Unbiased Watermarking \citep{hu2024unbiased} on Meta's OPT model
using a sample of the C4 dataset \citep{stahlberg-kumar-2021-synthetic}. 
% and MarkLLM\footnote{\url{https://github.com/THU-BPM/MarkLLM}} framework \citep{pan2024markllmopensourcetoolkitllm}, 
Thus, for RQ1, we evaluate these algorithms on 
various adversarial attacks (e.g., \textit{paraphrasing} via LLaMA-3-instruct-8B and 
\textit{back translation} using Multilingual LLaMA) to test their resilience. 
Next, for RQ2, 
% We evaluated both detection metrics, such as AUC scores and confusion matrix analysis, and 
we adopt several linguistic features, including POS tagging, sentiment analysis, Levenshtein distances, and descriptive statistics to ensure that watermarking preserves the natural quality and stylistic integrity of the generated text. 
This approach extracts critical insights for the linguistic characteristics of each watermark technique. 
Finally, for RQ3, we
highlight the trade-offs between watermark robustness and text quality of the algorithms 
by performing a linear correlation test between the adversarial robustness of the watermarking algorithms (using predictions) 
and the watermarking text quality (using the linguistic features). 
These results in a heatmap of Pearson's correlation tests for each algorithm. 
% Furthermore, we analyzed the relationship between linguistic features and the robustness of each watermarking method. We extracted a range of linguistic attributes—including POS distributions, sentiment scores, descriptive distributions (such as word count and sentence count), and Levinshtein distances by character—and compared them to the AUC scores of each watermarking algorithm to identify correlations between linguistic structure and detection performance. 
Lastly, by integrating these linguistic insights with quantitative evaluations, we provided a holistic understanding of how watermarking techniques interact with textual characteristics, offering a foundation for optimizing future watermarking strategies.

\section{Related Work} 

\subsection{Watermarking for LLMs}
Recent studies have explored watermarking techniques for LLMs.
These techniques can be broadly categorized based on their integration stage: during logits generation, token sampling, or LLM training \citet{liu2024surveytextwatermarkingera}. 
Each method presents distinct advantages and limitations in terms of robustness, detectability, and resistance to adversarial attacks. 
The more popular watermarking technique modifies logits 
by subtly adjusting the probability distribution of token outputs. 
\citet{kirchenbauer2024watermarklargelanguagemodels} proposes a method that shifts the logits distribution to favor certain sequences, embedding statistical patterns in the generated text. This approach ensures minimal perceptual impact while allowing post-hoc detection through statistical analysis. 
Another watermarking technique, performed at the token sampling phase benefits from ease of implementation, as it does not require modifying the model itself \citet{christ2023undetectablewatermarkslanguagemodels}.
Other approaches include trigger-based watermarks \citet{gu2023watermarkingpretrainedlanguagemodels, xu2024hufumodalityagnositcwatermarkingpretrained, sun2022coprotectorprotectopensourcecode}, 
and embedding watermarking during model training 
\citet{liu2024surveytextwatermarkingera}. 

However, to evaluate the robustness of watermarking techniques 
for LLMs, \citet{kuditipudi2024robustdistortionfreewatermarkslanguage} addresses two main challenges -  vulnerability to watermark removal attacks and the loss of randomness due to fixed pseudo-random sequences. Additionally, 
\citet{zhang2024remarkllmrobustefficientwatermarking} evaluates
that adversarial robustness of watermarking to paraphrasing and text smoothing adversarial attacks, finding that such attacks can weaken the detectability of such watermarks.

% % They proposed using a much longer pseudo-random sequence and randomly selecting a starting point for each watermark insertion to reintroduce randomness. Additionally, they incorporated a soft edit distance (Levenshtein distance) during watermark detection, which enhances robustness against attacks by improving the alignment between the text and the sequence.

% Another approach to watermarking involves embedding watermarks directly during model training. While watermarking during the logits generation and token sampling stages is effective in inference, these methods are not suitable for open-source LLMs, as watermark code added after the logits output can be easily removed. For open-source models, embedding watermarks into the LLM's parameters during training is necessary \citep{liu2024surveytextwatermarkingera}. This can be done through trigger-based watermarks, which activate for specific inputs, or global watermarks, which apply across all inputs. \citep{gu2023watermarkingpretrainedlanguagemodels, xu2024hufumodalityagnositcwatermarkingpretrained, sun2022coprotectorprotectopensourcecode}  

% However, studies such as \citet{zhang2024remarkllmrobustefficientwatermarking} indicate that adversarial techniques, including paraphrasing and text smoothing, can weaken the detectability of such watermarks.

\subsection{Linguistic Evaluations}
Several studies report that well-designed watermarks incur negligible quality loss by standard metrics. For example, \citet{nemecek2025topicbasedwatermarkslargelanguage} evaluate fluency by computing the perplexity of watermarked outputs under a large oracle model (LLaMA-3.1-8B). Their topic-guided watermark (TBW) yields perplexities nearly identical to unwatermarked text and significantly lower than other schemes such as Unigram \citep{zhaoprovable} or SynthID \citep{dathathri2024scalable}. Similarly, \citet{liu2024adaptivetextwatermarklarge} use LLaMA-13B to measure perplexity and report negligible impact on text quality; their watermarked outputs match the perplexities of both human-written and baseline model outputs, outperforming classic KGW variants. \citet{huo2024tokenspecificwatermarkingenhanceddetectability} also demonstrate that their token-specific watermark produces outputs with fluency and coherence comparable to the baselines, with oracle-model perplexity indistinguishable from clean text.

In contrast, \citet{singh2023newevaluationmetricscapture} argue that perplexity alone may obscure more nuanced degradations in linguistic quality. Using an LLM-based evaluator, they find that some watermarking schemes reduce coherence and response depth, revealing trade-offs not captured by traditional metrics. These findings highlight the need for multi-dimensional evaluations that account for both statistical and semantic properties of watermarked text.

\begin{table*}[!htb]
\footnotesize
    \centering
    \begin{tabular}{p{2cm} p{12cm}}
        \toprule
        \textbf{Technique} & \textbf{Description} \\
        \midrule
        KGW & 
        Introduced by \citet{kirchenbauer2024watermarklargelanguagemodels},
     this algorithm splits the model's vocabulary into a green list and a red list using a pseudorandom function based on the preceding token context. During text generation, a small positive bias is added to the logits of tokens belonging to the green list, thereby increasing their sampling probability. Detection is performed by computing a statistical z-score from the observed proportion of green tokens relative to the expected value. \\

     SIR & 
     Proposed by \citet{liu2024semanticinvariantrobustwatermark}, embeds watermarks based on semantic content rather than token identities. 
    SIR uses an embedding model (e.g., BERT) to extract semantic embeddings from prior tokens, which are then processed by a trainable watermark model to generate watermark logits. 
    This design resists semantically invariant transformations (e.g., paraphrasing) while maintaining security against statistical analysis. Detection relies on a z-score test to identify significant deviations in average watermark logits. \\

    Unbiased &  
    Proposed by \citet{hu2024unbiased} is designed to embed detectable signals in language model outputs without altering the model’s output distribution or degrading text quality. Unlike traditional methods that bias token probabilities, this approach uses unbiased reweighting functions.
    Watermark codes are generated from the context and a secret key using a pseudorandom function, making each watermark unique and undetectable without the key. 
    Detection is performed using a log-likelihood ratio test or its robust maximin variant, enabling reliable watermark identification even under text editing. \\
    
    EWD &  
    Proposed by \citet{lu2024entropybasedtextwatermarkingdetection}, enhances watermark detection in low-entropy scenarios like code generation. Traditional methods treat all tokens equally, causing low-entropy tokens which are highly predictable and difficult to modify. 
    EWD addresses this by assigning higher weights to high-entropy tokens, which are easier to watermark, and lower weights to low-entropy tokens, reducing their influence on the detection score. 
    The algorithm is training-free, automatically computes token weights from model logits, and applies a weighted sum to calculate the watermark score. \\

        \bottomrule
    \end{tabular}
    \caption{Description of the four Watermarking techniques we select for our study}
    \label{tab:water_tech}
\end{table*}

\begin{figure}%[!htb]
    \centering
    \includegraphics[width=1\linewidth]{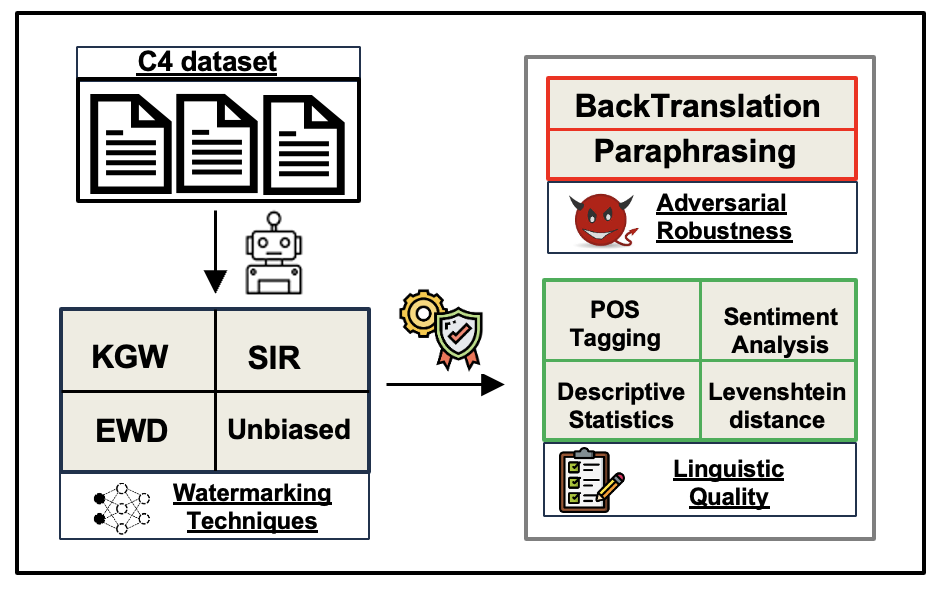}
    \caption{Illustration of the research pipeline}
    \label{fig:framework}
\end{figure}

\section{Problem Definition}
% As the quality of LLM-generated texts continues to become near indistinguishable from human-written texts, 
% our society will have to find novel solutions to perform for accurate authorship attribution of these 
% text generators. 
% Given this phenomenon, we believe that watermarking these LLMs will provide us with the ability to always perform accurate authorship attribution for LLM-generated texts. 
% However, before these watermarking techniques are adopted by LLM creators, we need to evaluate their ability to be robust to adversarial perturbations that aim to remove the watermarking signals within these texts, thus confusing the authorship, as well as evaluate the quality of these watermarked LLM-generated texts. 
To comprehensively investigate the strength of watermarking to combat authorship attribution of LLM-generated texts, 
we select four watermarking techniques - KGW, SIR, Unbiased, and EWD. See Table \ref{tab:water_tech} for a description of the watermarking techniques. This study yields three Research Questions; see below for detailed descriptions of the RQs.

\subsection{RQ1: Adversarial Attacks} 
We evaluate the robustness of our watermark detection method against common adversarial attacks aimed at removing or obfuscating watermarks. 

\begin{itemize}[noitemsep]
    \item \textbf{BackTranslation:}
    Backtranslation involves translating the watermarked text into an intermediate pivot language and then back into the original language \citep{he2024watermarkssurvivetranslationcrosslingual}. This process introduces subtle linguistic variations, disrupting the watermark while preserving semantic content. 

    \item \textbf{Paraphrasing:}
    Paraphrasing involves rewriting watermarked text while preserving its original meaning but altering lexical choices and sentence structures. This method exploits the vulnerability of AI-generated text detectors, as rewording can obscure watermark patterns \citep{krishna2023paraphrasingevadesdetectorsaigenerated}. 
 
\end{itemize}

\subsection{RQ2: Text Quality}
To assess the quality of the watermarked vs. unwatermarked LLM-generated texts, we use linguistic features to capture 
coherence, sentiment consistency, etc. See features:

\begin{itemize}[noitemsep]
    \item \textbf{POS tagging}:
    Each word in the text is labeled with its syntactic category (e.g., noun, verb, adjective), allowing us to quantify the frequency of each part of speech (POS). To measure whether watermarking altered grammatical patterns, we compared the POS distributions of watermarked and unwatermarked texts. 
    
    \item \textbf{Sentiment analysis}:
    To examine whether watermarking skews the emotional tone of the generated text, we applied a pre-trained RoBERTa-based sentiment classifier \citep{barbieri-etal-2020-tweeteval} to categorize outputs as positive, negative, or neutral. We compared the sentiment distributions across all watermarking methods and the original unwatermarked text. 
    
    \item \textbf{Descriptive statistics}:
    We calculated several descriptive text metrics to capture structural changes across watermarked and unwatermarked outputs, including the average sentence length, number of sentences, word length, number of words, and total text length. 
    %For each metric, we computed the mean across all outputs for each watermarking technique, comparing watermarked and unwatermarked samples.

    \item \textbf{Levenshtein distance}:
    To measure the degree of textual alteration, we calculated the Levenshtein distance between the original and watermarked outputs. This metric represents the minimum number of word-level edits (insertions, deletions, or substitutions) needed to transform one text into another.
\end{itemize}

Finally, see Figure \ref{fig:framework} for a visual representation of this study. 

\subsection{RQ3: Text Quality vs. Predictions}
% \adaku{I need to revisit this section}
Using Pearson's correlation test, we evaluate the linear relationship between the text quality and the correct \& incorrect predictions of the watermarked text. 
Thus, we compare the average counts/scores of the linguistic features with the classifier predictions. 
We aim to observe which linguistic metrics correlate more with predictions, thus revealing 
why some watermarking techniques may be more or less robust to certain adversarial attacks.

\begin{figure*}[!htb]
    \centering

    \begin{subfigure}[b]{0.32\textwidth}
        \includegraphics[width=\linewidth]{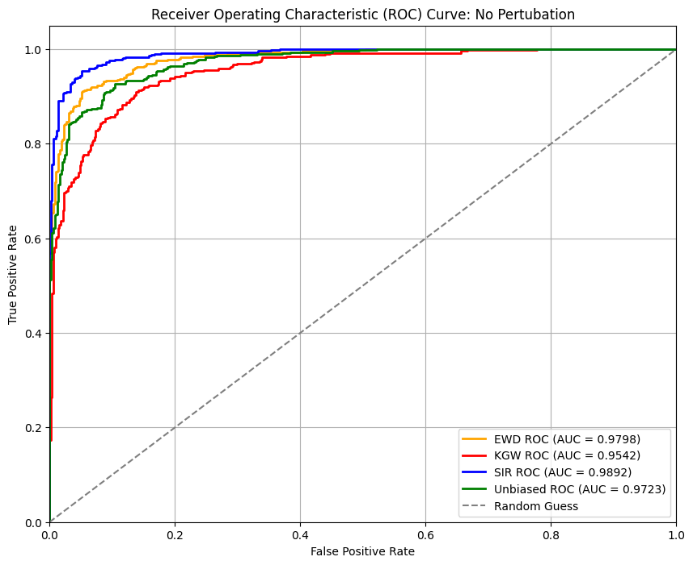}
        \caption{No perturbation}
    \end{subfigure}
    \hfill
    \begin{subfigure}[b]{0.323\textwidth}
        \includegraphics[width=\linewidth]{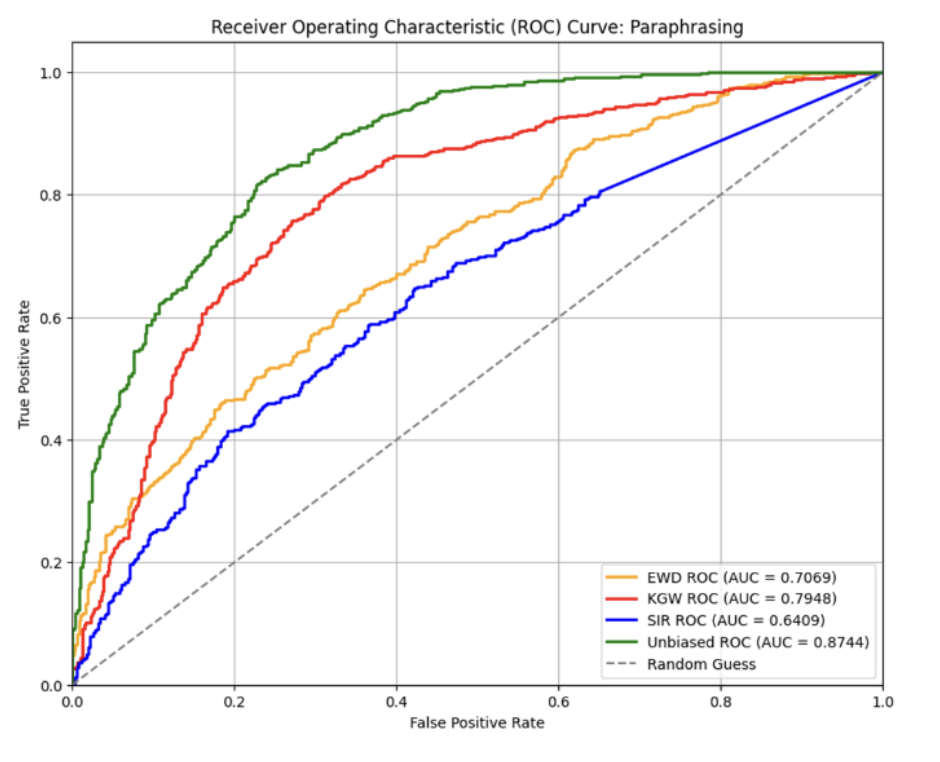}
        \caption{Paraphrasing}
    \end{subfigure}
    \hfill
    \begin{subfigure}[b]{0.32\textwidth}
        \includegraphics[width=\linewidth]{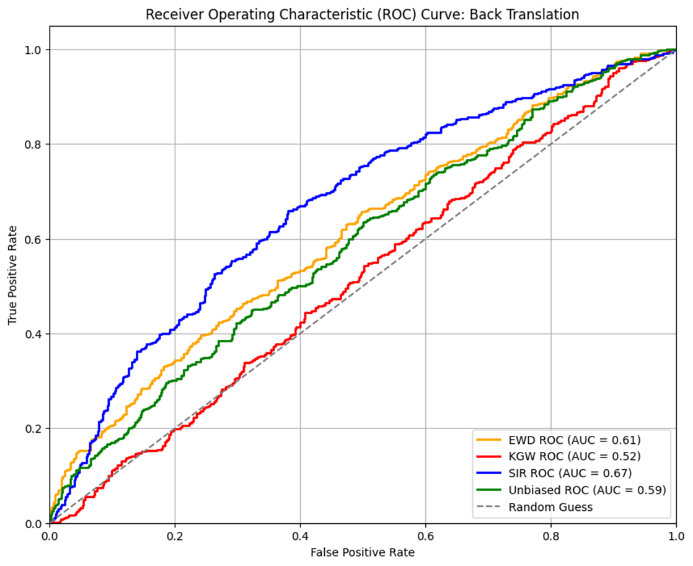}
        \caption{Backtranslation}
    \end{subfigure}

    \caption{ROC curves for EWD, KGW, SIR, and Unbiased under various attack methods: No perturbation, backtranslation, or paraphrase. }
    \label{fig:roc_curves}
\end{figure*}

\section{Methodology}

\subsection{Data Description}

We uses a subset of the Colossal Clean Crawled Corpus (C4) dataset,
which is a large-scale English-language text dataset \citet{JMLR:v21:20-074}. 
%as part of the T5 (Text-to-Text Transfer Transformer) model. 
C4 is constructed from the Common Crawl, a massive publicly available web scrape that undergoes extensive preprocessing to ensure high-quality text. The filtering process removes non-English content, duplicates, boilerplate text, and toxic language, resulting in a cleaner, more representative dataset of naturally occurring web text. The final processed C4 dataset consists of approximately 10,223 prompts from diverse text sources, including news articles, Wikipedia pages, blogs, and discussion forums. The selected passages ranged in length from 50 to 500 words, ensuring a mix of short and long-form text. 

\subsection{Evaluation}

% \paragraph{Performance:}
In our study, we leverage the MarkLLM pipeline\footnote{\url{https://github.com/THU-BPM/MarkLLM}} \citep{pan2024markllmopensourcetoolkitllm} to assess the watermarking techniques. 
% the resilience of watermarking algorithms against adversarial transformations. 
The pipeline encompasses the generation of the watermarked techniques employing Facebook OPT-1.3B \citep{zhang2022opt}. We then introduced adversarial modifications to the watermarked text. 
Two primary attack methods were employed: 
paraphrasing using LLaMA-3-8 B-Instruct \citep{llama3modelcard} which rephrased sentences while preserving meaning, 
and backtranslation via Multilingual LLaMA-3-8B \citep{devine2024tagengo} where text was translated from English to
French and then back to English.

Following these transformations, we deploy the \textit{WatermarkedTextDetectionPipeline} and \textit{UnWatermarkedTextDetectionPipeline} to analyze the detectability of the watermark in the unperturbed, paraphrased, and backtranslated texts. These pipelines employ specialized text editors, such as the \textit{TruncatePromptTextEditor}, to pre-process the text and extract detection confidence scores.

% To evaluate detection, we utilize the DynamicThresholdSuccessRateCalculator as we change the threshold, which calculates critical performance metrics, including the True Positive Rate (TPR), True Negative Rate (TNR), False Positive Rate (FPR), False Negative Rate (FNR), and F1 scores.

We employ the Area Under the Curve (AUC) of the Receiver Operating Characteristic (ROC) curve to obtain performance scores. AUC quantifies the likelihood that a randomly selected watermarked sample will be assigned a higher detection confidence score than a randomly chosen unwatermarked sample. A higher AUC value, approaching 1.0, indicates superior discriminatory performance, while a value of 0.5 suggests no distinction beyond random chance.
Additionally, we record results with ROC curves. 

Next, for the linguistic features, we record the average count of the selected sub-features within the list of 
features. 

% We also evaluate the linguistic features of the generated text.

% First is Parts of Speech (POS) Tagging, where each word in the text was labeled with its syntactic category (e.g., noun, verb, adjective), allowing us to quantify the frequency of each part of speech. To measure whether watermarking altered grammatical patterns, we compared the POS distributions of watermarked and unwatermarked texts. 

% To examine whether watermarking skews the emotional tone of the generated text, we applied a pre-trained RoBERTa-based sentiment \citep{barbieri-etal-2020-tweeteval} classifier to categorize outputs as positive, negative, or neutral. We compared the sentiment distributions across all watermarking methods and the original unwatermarked text. 

% We also calculated several descriptive text metrics to capture structural changes across watermarked and unwatermarked outputs, including the average sentence length, average number of sentences, average word length, average number of words, and total text length.For each metric, we computed the mean across all outputs for each watermarking technique, comparing watermarked and unwatermarked samples.

% To measure the degree of textual alteration, we calculated the Levenshtein distance between original and watermarked outputs. This metric represents the minimum number of character-level edits (insertions, deletions, or substitutions) needed to transform one text into another. 

\section{Results}
\subsection{RQ1: Adversarial Robustness}
As shown in Figure~\ref{fig:roc_curves}, under the no-perturbation condition, all four watermarking techniques achieved near-perfect detection performance. The SIR method achieved the highest AUC of 0.99, followed closely by EWD with 0.98, Unbiased with 0.97, and KGW with 0.95. 
% This indicates that all watermarking methods are highly effective at distinguishing watermarked from non-watermarked text when the output remains unaltered, with minimal degradation in classifier performance.

However, once paraphrasing was introduced as an adversarial attack, there was a noticeable drop in performance across all techniques. The Unbiased watermark retained the highest robustness, with an AUC of 0.87. KGW followed with an AUC of 0.8, and EWD with 0.71. 
SIR achieved the worst performance with AUC of 0.64. 
In addition, the ROC curves further illustrate this trend, with the Unbiased method maintaining a strong separation between true and false positives, whereas SIR curves approached random guessing more closely.

Under the back translation condition, performance deteriorated even further. The SIR technique showed relatively higher resilience in this context, achieving an AUC of 0.67, outperforming the other methods. EWD followed at 0.61, while Unbiased and KGW both exhibited diminished detection capabilities with AUCs of 0.59 and 0.52, respectively. 
% The ROC curves reflect this decrease in separability, highlighting the effectiveness of back translation as an evasion strategy.

% Comparing all conditions side by side in Figure~\ref{fig:auc_bar}, the Unbiased watermarking method offers the best trade-off between performance and resilience across both paraphrasing and back translation. SIR, while strong in clean text and relatively robust to back translation, is more susceptible to paraphrasing. KGW and EWD maintain moderate performance under paraphrasing but are less robust under back translation.

\begin{figure*}[!htb]
    \centering
    \includegraphics[width=0.9\textwidth]{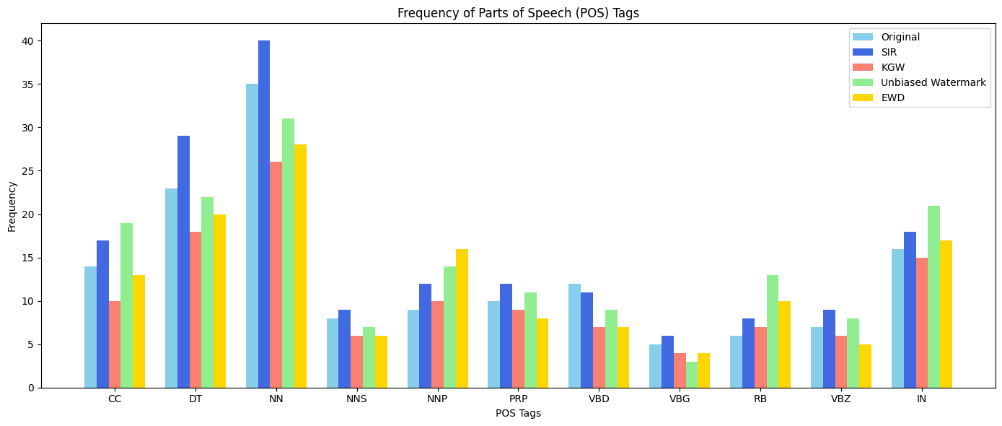}
    \caption{Bar graph distribution of POS tags colored by watermarking methods.}
    \label{fig:pos_distribution}
\end{figure*}

\begin{figure*}[!htb]
    \centering
    \includegraphics[width=0.8\textwidth]{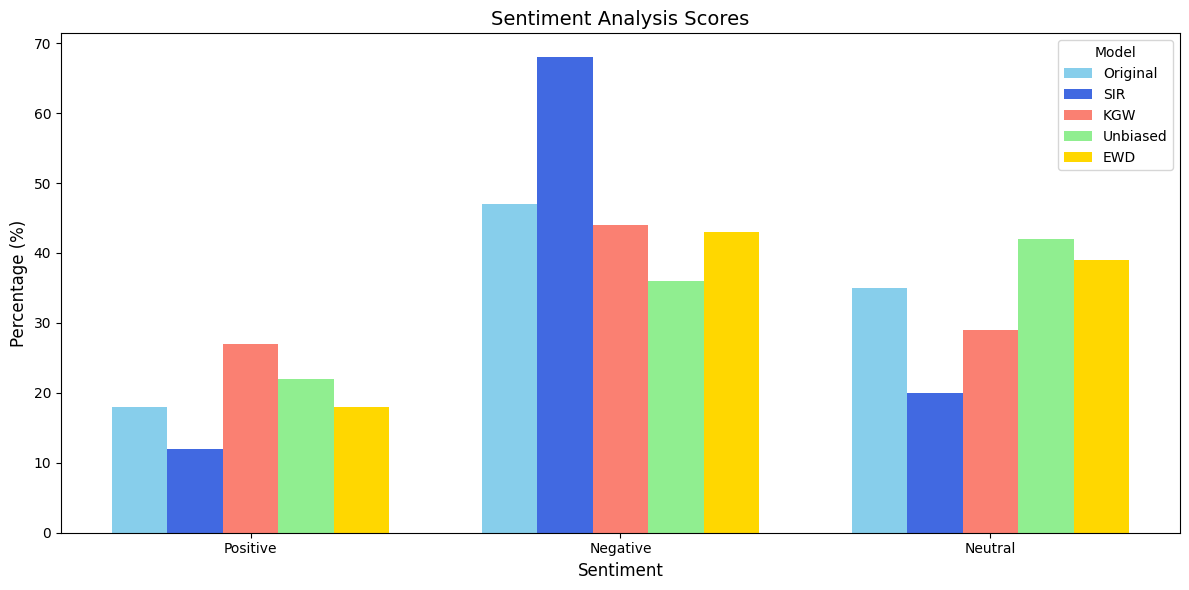}
    \caption{Sentiment distribution across watermarking methods.}
    \label{fig:sentiment_distribution}
\end{figure*}

\begin{table}[!htb]
    \centering
    \caption{Part-of-Speech (POS) Tag Counts by Watermarking Method}
    \label{tab:pos_table}
    \begin{adjustbox}{width=0.5\textwidth}
    \begin{tabular}{lrrrrr}
        \toprule
        \textbf{POS Tag} & \textbf{Original} & \textbf{SIR} & \textbf{KGW} & \textbf{Unbiased} & \textbf{EWD} \\
        \midrule
        NN   & 35 & 40 & 26 & 31 & 28 \\
        DT   & 23 & 29 & 18 & 22 & 20 \\
        IN   & 16 & 18 & 15 & 21 & 17 \\
        PRP  & 10 & 12 & 9  & 11 & 8  \\
        VBD  & 12 & 8  & 7  & 6  & 7  \\
        VBG  & 5  & 4  & 3  & 3  & 3  \\
        RB   & 6  & 8  & 7  & 13 & 10 \\
        NNP  & 9  & 12 & 11 & 14 & 16 \\
        \bottomrule
    \end{tabular}
    \end{adjustbox}
\end{table}

\subsection{RQ2: Text quality assessment}

\subsubsection{Part-of-Speech (POS) Tag Distributions}

Results are shown in Table~\ref{tab:pos_table} and Figure~\ref{fig:pos_distribution}. 
Examination of POS tag distributions reveals notable syntactic deviations introduced by each watermarking method relative to the original text.

Across the five scenarios (i.e., Original, SIR, KGW, Unbiased Watermark, and EWD), noun usage (NN) consistently dominates, although frequencies differ significantly. 
The SIR watermark increases the singular noun frequency from 35 to 40, while KGW reduces it to 26. 
EWD with 28 and Unbiased with 31 sit closer to the original.
%, suggesting moderate perturbation of nominal structures. 

For determiners (DT), SIR again increases the count from 23 to 29, whereas KGW and EWD reduce the number of DT to 18 and 20, respectively. 
% KGW with 18 and EWD with 20 reduce the amount of DT, possibly opting for more compact phrasing. 
A similar pattern holds for prepositions and subordinating conjunctions IN, where SIR with 18 and Unbiased with 21 mirror or exceed the original usage of 16, but KGW with 15 and EWD with 17. 
%lag slightly behind.

Pronouns (PRP) and verb forms such as VBD and VBG exhibit less consistent shifts. SIR and Unbiased watermarking show small increases in PRP of 12 and 11 vs. the original 10, while KGW with 9 and EWD with 8 reflect decreases. Verb usage drops across all watermarking methods.
% with VBD falling from 12 originally to 7 for KGW, EWD and VBG from 5 to as low as 3 Unbiased.

Adverbs (RB) provide one of the most telling contrasts. The Unbiased method nearly doubles adverb usage from 6 to 13, while SIR and KGW slightly elevate or match original levels to 8 and 7, respectively, while EWD exhibits conservative increases with 10 counts. Proper nouns (NNP) vary across strategies, with SIR 12, Unbiased 14, and EWD 16 exceeding the original count 9.

% Among all methods, the Unbiased watermark most closely resembles the Original in POS tag balance, indicating its effectiveness in preserving syntactic fidelity. SIR induces the most substantial perturbations, particularly in nominal and functional categories. KGW shows the lowest frequencies, suggesting a minimalistic approach, while EWD falls in between.

\begin{table}[!htb]
    \centering
    \caption{Sentiment Label Proportions by Watermarking Method (\%)}
    \label{tab:sentiment_table}
    \begin{tabular}{lccc}
        \toprule
        Method & Negative\% & Neutral\% & Positive\% \\
        \midrule
        Original & 47 & 35 & 18 \\
        SIR & 68 & 20 & 12 \\
        KGW & 44 & 29 & 27 \\
        Unbiased & 36 & 42 & 22 \\
        EWD & 43 & 39 & 18 \\
        \bottomrule
    \end{tabular}
\end{table}

\subsubsection{Sentiment Analysis}
The distribution of sentiment labels (i.e., Positive, Negative, and Neutral) is illustrated in Figure~\ref{fig:sentiment_distribution} and summarized in Table~\ref{tab:sentiment_table}.
The Original texts' sentiment profile contains
47\% Negative, 35\% Neutral, and 18\% Positive. 
SIR watermarking introduces a skew toward Negative sentiment with 68\%, a drop in Neutral and Positive to 20\% and 12\%, respectively. 
% This suggests that SIR may embed information by amplifying pessimistic or critical phrasing.
KGW watermarking sentiment profile has 44\% Negative, 29\% Neutral, and 27\% Positive, exhibiting the highest Positive rate among all the methods. 
Unbiased watermarking yields the most sentimentally neutral outputs (42\% Neutral, 36\% Negative, 22\% Positive), also closely mirroring the Original distribution. 
Finally, EWD watermarking shows 43\% Negative, 39\% Neutral, and 18\% Positive.
% presenting a middle ground between Unbiased and SIR in emotional tone.

\subsubsection{Levenshtein Distance}
\begin{figure}
    \centering
    \includegraphics[width=0.45\textwidth]{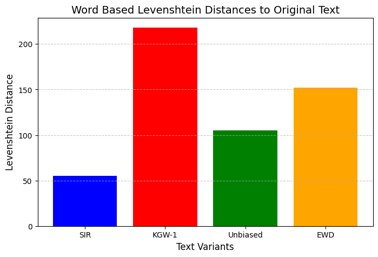}
    \caption{Levenshtein distance from the original text across watermarking methods.}
    \label{fig:levenshtein}
\end{figure}

As shown in Figure~\ref{fig:levenshtein}, the KGW method resulted in the highest Levenshtein distance of 823, indicating the greatest divergence from the original text. EWD followed with a distance of 600, Unbiased with 375, and SIR with the smallest distance of 127, implying minimal alteration.

\subsubsection{Text Statistics}
Table~\ref{tab:text_stats} provides additional insight. 
KGW yielded the lowest average sentence length of 19.57 words and the shortest word length of 4.17 characters. 
The Unbiased method also reduced sentence and word length. 
SIR maintained linguistic features closest to the original, while EWD introduced moderate structural changes.

\begin{table*}
    \centering
    \caption{Text Statistics by Watermarking Method: Mean of count}
    \label{tab:text_stats}
    \begin{tabular}{lcccc}
        \toprule
        Method & Sent. Length (words) & Word Length (chars) & Text Length (chars) & Levenshtein Distance \\
        \midrule
        Original & 24.50 & 4.44 & 1403 & -- \\
        SIR & 24.38 & 4.42 & 1370 & 127 \\
        KGW & 19.57 & 4.17 & 1281 & 823 \\
        Unbiased & 21.80 & 4.31 & 1345 & 375 \\
        EWD & 22.20 & 4.28 & 1328 & 600 \\
        \bottomrule
    \end{tabular}
\end{table*}

% \begin{figure}[!htb]
%     \centering
%     \includegraphics[width=\linewidth]{auc_bar.jpg}
%     \caption{Bar graph distribution of AUC scores colored by watermarking technique}
%     \label{fig:auc_bar}
% \end{figure}

\subsection{RQ3: Text Quality vs. Predictions}

\begin{figure}[!htb]
    \centering
    \includegraphics[width=0.45\textwidth]{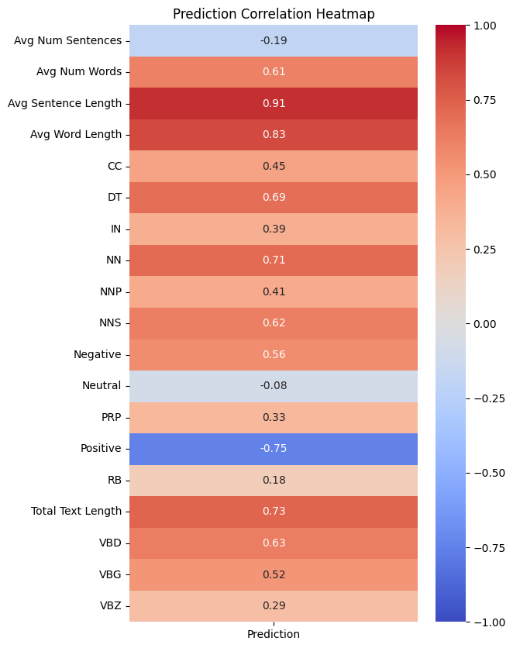}
    \caption{Pearson Correlation Coefficients of Linguistic features vs. Predictions}
    \label{fig:Pearson}
\end{figure}

Figure~\ref{fig:Pearson} presents a heatmap of the correlation values between the predictions and the linguistic features.
These scores are the combination for all the four watermarking techniques. 
The predictions (i.e., correct \& incorrect) correlate most strongly with average sentence length ($r = 0.91$) and average word length ($r = 0.83$), followed by total text length ($r = 0.73$), noun frequencies (NN, $r = 0.71$), determiners (DT, $r = 0.69$), and past tense verbs (VBD, $r = 0.63$). Other notable positive correlations include plural nouns (NNS, $r = 0.62$), and average number of words ($r = 0.61$). Other POS tags, such as proper noun frequency (NNP), prepositions (IN), verbs in the present participle (VBG), coordinating conjunctions (CC), adverbs (RB), and base verbs (VBZ), display moderate to weak positive correlations ranging from $r = 0.29$ to $r = 0.52$.
Positive sentiment shows a strong negative correlation with AUC ($r = -0.75$), while neutral sentiment ($r = -0.08$) exhibits a weak negative correlation, and negative sentiment ($r = 0.56$) exhibits a moderate positive correlation. 

\section{Discussion}

\subsection{SIR and Unbiased Watermarking Better Preserve LLM Style and Fluency}

Beyond adversarial robustness, a watermarking method must minimize disruption to linguistic quality to ensure naturalness and avoid detection by users. Our linguistic analyses suggest that SIR and Unbiased watermarking methods strike the best balance between fidelity to the original style and effective embedding.

SIR watermarking exhibits the smallest Levenshtein distance from the original text (127), indicating minimal surface-level alteration. While its POS distribution (Table~\ref{tab:pos_table}) deviates from the original more than Unbiased, it retains structural integrity in sentence length and word complexity (Table~\ref{tab:text_stats}). Conversely, the Unbiased watermark excels in maintaining syntactic and stylistic norms of the base language model, with POS tag distributions and sentiment profiles (Table~\ref{tab:sentiment_table}) closely aligned with the original. For instance, its sentiment skew remains centered with the highest neutral proportion (42\%) and modest shifts in noun and adverb usage.

By contrast, KGW and EWD introduce more pronounced stylistic distortions. KGW significantly reduces sentence length, alters POS tag distributions (e.g., lower noun and determiner counts), and has the highest Levenshtein distance (823), suggesting a more aggressive rewriting strategy. EWD occupies a middle ground, inducing moderate perturbations while maintaining some degree of syntactic resemblance to the original.

\subsection{Back Translation More Effectively Strips Watermarking Signatures Than Paraphrasing}
 
Our results suggest that back translation is a more effective evasion strategy than paraphrasing. While paraphrasing tends to rephrase sentences within the same language and thus preserve some latent structural watermarking features, back translation introduces cross-lingual transformations that often eliminate low-level syntactic and lexical markers. This is especially evident in the Unbiased and KGW methods, where AUC scores dropped precipitously under back translation (to 0.59 and 0.52, respectively) compared to paraphrasing. These findings are consistent with prior research \citep{he2024watermarkssurvivetranslationcrosslingual} suggesting that multilingual rewriting significantly reduces stylometric cues, effectively weakening watermark-related patterns—while maintaining the original semantic content.

\subsection{Linguistic Quality Correlates with Adversarial Robustness}
We observe from the Pearson correlation results (Figure~\ref{fig:Pearson}) 
that certain linguistic properties highlight what is watermarked vs. unwatermarked. 
Specifically, we observe that predictions are positively correlated with structural features such as average sentence length ($r = 0.91$), average word length ($r = 0.83$), and the presence of determiners ($r = 0.69$), nouns ($r = 0.71$), and past-tense verbs ($r = 0.63$). 
These patterns suggest that longer, syntactically dense sentences contribute to more resilient watermarking as 
observed from other linguistic analysis on the different watermarking techniques, including the performance of these techniques under adversarial constraints. 
Conversely, texts with more positive sentiment are significantly less robust ($r = -0.75$), 
which is confirmed by KGW achieving the highest positive sentiment (27\%), which led to the 
significant decline in performance when perturbed with the back translation attack (95\% $\to$ 52\%). 
This indicates that positive emotionally charged or affirming language may be more vulnerable to adversarial perturbations. The lack of strong correlation for neutral sentiment ($r = -0.08$) and part-of-speech categories such as adverbs ($r = 0.13$) suggests that not all stylistic features are equally predictive of watermark survivability. 
Overall, linguistic richness appears to support stronger watermark retention under attack, while stylistic simplicity or emotional positivity may hinder it.

% reveal that specific linguistic properties of the watermarked text are closely tied to its correct or incorrect predictions. 
% Predictions are positively correlated with structural features such as average sentence length ($r = 0.91$), average word length ($r = 0.83$), and the presence of determiners ($r = 0.69$), nouns ($r = 0.71$), and past-tense verbs ($r = 0.63$). 
% These patterns suggest that longer, syntactically dense sentences contribute to more resilient watermarking.

% Conversely, texts with more positive sentiment are significantly less robust ($r = -0.75$), indicating that emotionally charged or affirming language may be more vulnerable to adversarial perturbations. The lack of strong correlation for neutral sentiment ($r = -0.08$) and part-of-speech categories such as adverbs ($r = 0.13$) suggests that not all stylistic features are equally predictive of watermark survivability. Overall, linguistic richness appears to support stronger watermark retention under attack, while stylistic simplicity or emotional positivity may hinder it.

\section{Conclusion}
This study provides a comparative analysis of neural watermarking methods under adversarial conditions. Among the approaches tested, the Unbiased watermark strikes the best balance between detectability and robustness, performing consistently across both paraphrasing and back translation attacks. SIR performs well on clean text but is more vulnerable to paraphrasing, while KGW and EWD show moderate resilience, especially under back translation. 
Our analysis of linguistic features reveals that certain characteristics, such as longer sentences, longer words, and greater use of nouns, are positively associated with better watermark robustness. 
In contrast, a more positive sentiment is linked to decreased resilience. 
These findings suggest that the linguistic makeup of a text plays a key role in how well watermarking holds up under attack, and they point to useful design considerations for future methods.

% \newpage

\section{Limitations}
\begin{itemize}
    \item \textbf{Dataset Constraints:} 
    All linguistic evaluations were conducted exclusively on a cleaned subset of the English portion of the C4 dataset. This limits generalizability, as watermark robustness and detectability may vary across domains such as code generation, dialogue systems, multilingual corpora, and low-resource or noisy text environments.

    \item \textbf{Single LLM Architecture:}
    All experiments were conducted using OPT-13B, which differs in training data, architecture, and decoding behavior from other widely used models. Further evaluation on models like LLaMA-3, Mistral, Deepseek, and Phi could reveal model-specific vulnerabilities or sensitivities to watermarking techniques.

    \item \textbf{Limited Watermarking Coverage in MarkLLM:}
    While MarkLLM implements several representative watermarking strategies, it does not encompass all existing techniques. 
    
    \item \textbf{Restricted Adversarial Scope:}
    The set of adversarial attacks explored was limited. Techniques such as decoding manipulation (e.g., temperature tuning, nucleus sampling), prompt injection, and adversarial fine-tuning were not evaluated, leaving gaps in comprehensively evaluating watermark robustness under adversarial pressure.

    \item \textbf{Underexplored Linguistic and Stylistic Features:}
    Our analysis focused on surface-level token and frequency distributions. Other linguistic signals—including syntactic depth, discourse structure, semantic coherence, and stylistic traits like formality, hedging, or lexical diversity—were not analyzed, which can lend further insight into watermarking robustness and linguistic features.

\end{itemize}

\section{Ethical Statement}
This study investigates the robustness of large language model (LLM) watermarking techniques, with the goal of advancing the reliability and security of such methods. While the techniques explored in this work could inform attempts to circumvent watermarking mechanisms, our primary intent is to surface vulnerabilities that can be proactively addressed. By examining and disclosing these limitations, we aim to support the development of more robust and trustworthy watermarking strategies. We believe that the benefits of this transparency, including improvements in detection and policy alignment, outweigh the potential risks.

% To evaluate the robustness of watermarking techniques for LLMs, 
% we assess the adversarial robustness and text quality of different 
% watermarking techniques. We understand that results from this 
% study can potentially be used by malicious actors to strip 
% watermarking signatures from LLM-generated texts, thus 
% evading detection. 
% However, we believe that our results can be used to improve watermarking technique, thus thwarting any evasion techniques by 
% malicious actors. Finally, this implies that 
% the benefits outweigh the risk of this research study. 

% \section*{Acknowledgments}
% We thank Dr. Adaku Uchendu and Dr. Ana Smith at MIT Lincoln Laboratory for their guidance and feedback throughout the development of this research. 

\bibliography{custom}

\appendix

% \section{Example Appendix}
% \label{sec:appendix}

% This is an appendix.

\end{document}